# Beyond One-Time Validation: A Framework for Adaptive Validation of Prognostic and Diagnostic AI-based Medical Devices

Florian Hellmeier[1]*, Kay Brosien[1]*, Carsten Eickhoff[2], Alexander Meyer[1,3,4,5]

[1] x-cardiac GmbH, Berlin, Germany
[2] Institute for Bioinformatics and Medical Informatics, University of Tübingen, Tübingen, Germany
[3] Department of Cardiothoracic and Vascular Surgery, Deutsches Herzzentrum der Charité, Berlin, Germany
[4] DZHK (German Centre for Cardiovascular Research), partner site Berlin, Berlin, Germany
[5] Berlin Institute for the Foundations of Learning and Data, Berlin, Germany
* These two authors contributed equally to this manuscript.
Corresponding author: Alexander Meyer, alexander.meyer@dhzc-charite.de

**Abstract**

Prognostic and diagnostic AI-based medical devices hold immense promise for advancing healthcare, yet their rapid development has outpaced the establishment of appropriate validation methods. Existing approaches often fall short in addressing the complexity of practically deploying these devices and ensuring their effective, continued operation in real-world settings.

Building on recent discussions around the validation of AI models in medicine and drawing from validation practices in other fields, a framework to address this gap is presented. It offers a structured, robust approach to validation that helps ensure device reliability across differing clinical environments.

The primary challenges to device performance upon deployment are discussed while highlighting the impact of changes related to individual healthcare institutions and operational processes. The presented framework emphasizes the importance of repeating validation and fine-tuning during deployment, aiming to mitigate these issues while being adaptable to challenges unforeseen during device development.

The framework is also positioned within the current US and EU regulatory landscapes, underscoring its practical viability and relevance considering regulatory requirements. Additionally, a practical example demonstrating potential benefits of the framework is presented. Lastly, guidance on assessing model performance is offered and the importance of involving clinical stakeholders in the validation and fine-tuning process is discussed.

## Introduction

In the rapidly evolving digital healthcare landscape, validation and regulation of artificial intelligence (AI) and machine learning models has emerged as a pivotal challenge for trustworthy and effective use of such solutions. Unlike conventional biomarkers or monitoring and imaging devices, which typically measure individual biochemical or biophysical quantities, AI-based medical devices operate on more abstract types of data. They often utilize and process multiple outputs from traditional tools as inputs. This article will focus on prognostic and diagnostic AI-based medical devices (AI-MDs), i.e. those that diagnose current or predict future health states or outcomes.

Classic laboratory values or imaging data provide straightforward and well-established pathways for validation, based on directly measurable quantities. In contrast, prognostic and diagnostic AI models can process and analyze these and other parameters to infer insights or predict outcomes, thus operating on a level that integrates and synthesizes information across various sources and formats. This additional abstraction layer not only complicates the development process but also amplifies the challenges associated with generalization and its confirmation through validation, necessitating novel approaches to confirm the accuracy, reliability, and clinical relevance of diagnoses or predictions generated by prognostic and diagnostic AI-MDs.

The validation of these devices demands a nuanced approach that accounts for the complexity and diversity of the data they process. It requires a framework that can evaluate the performance of these models across a wide spectrum of clinical scenarios and patient populations to ensure adequate model behavior. Such a framework needs to account for the dynamic nature of clinical data and ensure that, as deployment environments change and models encounter new information, they possess the capability to adapt accordingly.

We propose an operationalizable framework based on repeating local validation coupled with local model fine-tuning, building on recent discussions in the literature.[1–5] The framework was conceived through collaborative discussions among the authors, whose backgrounds span academic research, medical device manufacturing, and clinical practice, and was further refined through discussions with colleagues across these fields. The proposed framework is intended to instill confidence among clinical users, while simultaneously providing a way for medical device manufacturers to ensure reliable performance of their devices and satisfy regulatory demands.

## Challenges regarding the performance of prognostic and diagnostic AI-MDs

Good generalization, i.e. good performance on previously unseen data, is a desirable property for any predictive or diagnostic model. Unfortunately, achieving it has often proven difficult for clinical AI models, with studies from a variety of specialties demonstrating suboptimal model generalization.[6–12] The issue of generalization is also recognized by a substantial portion of clinicians as a major challenge to the adoption of AI-based solutions.[13]
An important reason why achieving good generalization is often difficult, is the influence of the context of medical care delivery on model performance. Key factors including healthcare institutions' standard operating procedures, medical resource availability, staff levels, expertise, and patient demographics are often not explicitly considered in model design, partly because these elements are challenging to quantify or operationalize. Not considering these aspects may lead to dataset shift negatively affecting model performance. For instance, a prognostic model trained on one patient cohort might be used on a slightly different patient cohort. There likely will be patient-related factors that a) influence patient outcome and b) are neither captured by the model nor can they be explained by factors that are. This cohort-related dataset shift may then affect model performance. Finlayson et al.

provide a detailed overview of different dataset shift types as well as practical recognition and mitigation strategies.[14]

Regarding the deployment of prognostic and diagnostic AI-MDs, we will focus on two subtypes of dataset shift, which will be referred to as operational and institutional shift. *Institutional shift* will refer to changes in the model input/output distributions or the input-output relationship, caused by *deploying the model to a new healthcare institution*. It, for example, includes the effect of changes in patient cohort, equipment, and procedures between healthcare institutions. *Operational shift* will refer to changes in the model input/output distributions or the input-output relationship, caused by *time-related changes within the same healthcare institution*. It, for example, includes the effect of changes to staff levels and expertise, procedures, and disease spectrum at a specific healthcare institution over time. Both types of shift can affect model performance in new deployment environments or as environments change over time, highlighting the interplay between model design and the dynamic nature of clinical settings.

Recognizing that poor generalization negatively affects model performance, a common approach is to address it during model development. This may include identifying causes, such as dataset shifts, and attempting to mitigate them through model design and training. Use of large foundation models, as defined by Bommasani et al., may improve overall robustness in the future but these models also still face difficulties compensating for operational shift.[15] Acknowledging the challenges encountered when focusing primarily on model generalization, we advocate for broadening our perspective and moving a step back from focusing solely on the underlying causes of unsatisfactory performance. By employing a root cause-agnostic framework to continuously identify and remedy performance issues in actual use, our approach expands the focus from mitigating these issues only during development to also addressing them during operational use. Such a framework is not only inherently safer by enabling continuous assessment of model performance, it also allows accounting for performance deviations caused by effects that were not anticipated during model development.

**The REVAFT framework**

As the deployment of prognostic and diagnostic AI models in healthcare progresses, the limitations of traditional validation approaches, e.g. external validation, become more apparent. A variety of existing guidelines and frameworks cover a substantial part of the prognostic and diagnostic AI-MD lifecycle. These include TRIPOD+AI for reporting clinical prediction models[16], DECIDE-AI for early stage decision support system evaluation[17], and comprehensive frameworks, such as the WHO's 2021 "Generating Evidence for Artificial Intelligence Based Medical Devices" framework[18]. While the former two act as reporting guidelines for specific stages of the medical device lifecycle, the latter provides a higher-level overview, indicates minimum standards for different product lifecycle phases, and references the other guidelines where appropriate. A point that is rather cursorily covered in current guidelines and frameworks is the question of how appropriate device performance can be maintained during product deployment after achieving it during initial deployment to a new institution.

In light of the problems of generalization and time-dependent changes of model performance, approaches that utilize local validation and/or retraining have been suggested in recent literature.[1–4] When local validation reveals performance gaps, site-specific modifications to the model can be made through fine-tuning, i.e. by further model training using data from the new deployment environment. This has been demonstrated to improve model performance in clinical applications, e.g. by Mohn et al. who demonstrate performance improvements through fine-tuning for a diagnostic use case based on chest radiographs.[19] Fine-tuning also has the potential to be sample-efficient and requires relatively little data to achieve relevant performance gains.[20] To be practically viable for medical device manufacturers, the fine-tuning process must be implemented in a way that avoids complete recertification of the AI-MD. Since the regulatory landscape around this issue is currently evolving, no universally applicable guidance on how to achieve this, can be given here. However, preliminary indications suggest that fine-tuning, within certain limits, may be achievable without necessitating recertification in the US and EU.[21,22]

We concur with the aforementioned views on the importance of continuous evaluation of model performance,[1–3] emphasizing the importance of remedial action, should performance degrade. The idea behind the proposed REVAFT (REpeating VAlidation and Fine-Tuning) framework is to provide a concrete way to implement a validation approach that combines repeating validation and fine-tuning. Fig. 1A contrasts it against classical validation approaches. REVAFT is intended to be practically feasible for researchers, healthcare institutions, and medical device manufacturers, while also providing model performance transparency for clinical users. It includes the following main steps:

1. **Establishment of Baseline:** Begin with a base model that has undergone, at minimum, validation against one external cohort representative of the specific use case to establish a baseline.
2. **Regulatory Certification:** Certify the model as a medical device, confirming its regulatory compliance and readiness. Integrate a process within the quality management system for the ongoing fine-tuning of the medical device.
3. **Real-World Performance Assessment:** Assess the model's performance in each deployment environment using available historical data to identify institutional shift and gauge the model's real-world effectiveness, e.g. using descriptive statistics.
4. **Deployment-Specific Fine-Tuning:** Whenever feasible, perform fine-tuning at the point of deployment to optimize performance in the specific deployment environment.
5. **Ongoing Validation and Adaptation:** Repeat validation and fine-tuning at regular intervals, based on the deployment context and the potential for operational shift. Implement surveillance measures to detect operational shift and trigger validation and fine-tuning. Additionally, perform validation and fine-tuning in response to changes in clinical infrastructure or procedures, or following safety events. Reassess regulatory permissibility of continued medical device deployment following each validation and fine-tuning episode.

Fig. 1B illustrates steps 3 to 5 of the proposed REVAFT framework. If model performance initially or during operational use of the model falls below set thresholds, the algorithm is adjusted by incorporating historical data, customizing it to better match local conditions and practices. This fine-tuning process benefits from the fact that prognostic and diagnostic

AI-MDs are already based on clinical data sources, making extraction of historical data for model fine-tuning often straightforward and non-disruptive. It is assumed that the ground truth data necessary for fine-tuning can be obtained from clinical data sources. For a prognostic and diagnostic AI-MDs this would be a clinically confirmed diagnosis, outcome or health state. It should be noted that the approach is only viable if the aforementioned ground truth data can be obtained with reasonable effort, ideally automatically,, as it is required each time an episode of repeating validation and fine-tuning ist triggered. If substantial manual processing is necessary, it limits the applicability of the approach. In our experience, fine-tuning can often be performed using local computing resources, since it is generally less computationally expensive than the initial development and training of the base model. Depending on the frequency of validation and fine-tuning, availability of a large enough dataset for validation and fine-tuning might pose an issue, particularly for rare outcomes. In this case, data augmentation techniques may be necessary to obtain sufficient data for fine-tuning, while retaining enough non-augmented data for validation.

Model performance during the repeating validations needs to be assessed in a suitable manner. Depending on the model in question, the information and performance metrics needed to assess model performance may vary. A collection of tools may be used to communicate validation results. Depending on the model, these may, for instance, include statistical metrics (e.g. sensitivity, specificity, positive and negative predictive value, F1 score), calibration plots and metrics, model cards, and/or a glossary, to ensure that all stakeholders can fully appreciate the presented information. For time-resolved prognostic and diagnostic AI models performance evaluation also needs to accommodate time-resolved outputs. For these models, performance should be evaluated at representative time points during patients' clinical courses and consistency of model performance over time should be assessed. Based on the model, the statistical and/or performance metrics mentioned above may be used in a time-resolved manner. Additionally, for prognostic models performance leading up to predicted events and the timing between the events occuring and the predictions should be evaluated. Assessing these aspects is necessary to obtain a comprehensive picture of the model's performance over time as well as its clinical usefulness at different time points and should be included in the validation results.

Regarding the responsibility for repeating validation and fine-tuning, performance assessment can either be performed by the deploying institution alone or in cooperation with the device manufacturer. Validation results should be communicated to key clinical stakeholders. If a practically relevant degradation of device performance is observed, the manufacturer then needs to be involved. The reason being that, while hospitals might be able to decide to stop using an AI-MD that performs subpar, they cannot, on their own, fine-tune the device to restore acceptable device performance. Such changes to the device not only have regulatory implications, they also require technical access and insights, that only the device manufacturer has.

To determine the optimal timing for repeating validation and fine-tuning, three categories of triggers should be considered. The first category is a fixed schedule, with the frequency tailored to the specific use case and an estimate of the timescale over which operational shifts might occur. During early deployment, this schedule could be set with higher frequency, with the option to deescalate it as more data is gathered on the pace of operational shift. The second category includes specific events, such as device-related safety events or changes to clinical infrastructure or procedures. These events are primarily identified by clinicians, who would then notify the device manufacturer. The third category consists of continuous monitoring methods for device inputs and outputs. If ground truth data for prognostic devices can be obtained automatically, their performance can be continuously evaluated. Similarly, approaches, such as those proposed by Koch et al.[23] can be used to detect significant operational shift.

The REVAFT framework contrasts with other validation and training approaches, including one-time external validation and the more dynamic but currently impractical online learning. Table 1 compares traditional validation approaches plus vigilance and post-market surveillance with REVAFT. One-time external validation suffers from the fact that its results are unlikely to be representative of model performance in individual deployment environments. Online learning, on the other hand, currently faces technical and regulatory challenges, notably with regard to product versioning and algorithm reliability. Reliability in this context referring to the ISO/IEC 22989:2022 definition, which requires consistent intended behavior.[24] These issues currently limit the application of online learning in prognostic and diagnostic AI-MDs, though future changes to the regulatory landscape may alter its viability. REVAFT employs batch learning which enables periodic updates to prognostic and diagnostic AI-MDs with discrete adjustments at specific intervals, ensuring clear documentation and regulatory compliance through version control. This method meets the validation processes required by US and EU regulatory authorities, which do not explicitly restrict algorithm fine-tuning but demand comprehensive validation comparable to the initial device validation.[21,22] One pathway through certification involves incorporating fine-tuning processes within the quality management system. This allows for all related activities to be systematically documented, and the outcomes to be substantiated with appropriate documents, such as checklists. However, this nuanced understanding necessitates clearer regulatory guidance, particularly in the EU, regarding the definition and implications of "significant changes" in the context of algorithm updates and fine-tuning.

Since REVAFT requires ongoing access to sensitive patient information, it necessitates stringent data protection measures to ensure that the handling of patient data during repeating validation and fine-tuning adheres to the same standards expected during normal deployment of medical devices. It is imperative that comprehensive data protection provisions are clearly defined and agreed upon by healthcare institutions deploying these devices and their manufacturers. Ideally, REVAFT can be implemented entirely on computing resources of the deploying institution, minimizing risks to data security.

**A case for the proposed framework**

To demonstrate the potential benefits of the REVAFT framework, it is worth looking at the example of an EHR-based commercial solution for the prediction of sepsis, which has previously sparked discussion regarding generalization and utility of sepsis prediction models.[10,25–29] Initially developed on data from three health systems,[10] its generalization to other healthcare institutions as well as assessments of clinical usefulness have been heterogeneous.[10,27,29] While some of the performance differences seen in the mentioned studies are likely related to their respective designs, there nevertheless seems to be substantial differences in predictive performances depending on the deploying institution, indicating that institutional shift could not be fully compensated by the model.

To the authors' knowledge, the predictive model provided little opportunity for fine-tuning at the time the previously mentioned studies were performed. It allowed calibration in the sense that the sepsis alert threshold could be set by the deploying institutions.[10] This does provide some control over positive and negative predictive value, sensitivity, specificity, and alert frequency. It is, however, insufficient for fine-tuning the model to a particular institution's patient cohort and standards of care. Apart from institutional shift, the prediction model has also been affected by operational shift, caused by the emergence of COVID-19.[14] The change in predictive performance caused by this operational shift ultimately required deactivation of the model in at least one large academic hospital[14] and led to a substantial increase in alerts per patient in a sample of 24 hospitals,[30] reducing its usefulness.

Additionally, at least two studies found poor timeliness of the model's predictions.[27,28] Besides the fact that timely initiation of treatment is a cornerstone of successful sepsis therapy, the model seems to have included antimicrobial orders in its score calculation.[27] Given that such orders indicate clinicians' awareness of the possibility of sepsis, the usefulness of alerts triggering afterwards is likely low. Beaulieu-Jones et al. have demonstrated that it is comparatively easy to build EHR-based models that seemingly perform well, while doing little more than implicitly capturing the judgment of clinicians through their actions.[31] The susceptibility of EHR-based AI models to this phenomenon varies, depending on whether the model inputs are routinely captured – such as vitals – or are only obtained in specific situations – such as blood culture and antimicrobial orders.

Both, the performance inconsistencies caused by institutional shift and the operational shift caused by the COVID-19 pandemic, demonstrate the importance of recurring assessment of model performance. While operational shift caused by a pandemic might not have been foreseeable during model development, the factors underlying institutional shift, e.g. institution-related changes to patient cohort or procedures, are common and well-known issues. In any case, both institutional and operational shift demonstrate that a proactive approach to model validation and fine-tuning is necessary to ensure that model performance is adequate at different institutions and at different times. Beyond the problem of dataset shift, the issue of prediction timeliness could have been recognized using time-resolved evaluation of model performance, as demonstrated by Schertz et al.[27] For prognostic models, it is important to assess how predictive performance is related to the time difference between the prediction being made and the actual event occurring. It follows that model performance should be evaluated in a time-resolved manner.

In summary, use of the proposed REVAFT framework might have helped mitigate the problems encountered during deployment of the aforementioned sepsis prediction model. Local validation could have helped identify suboptimal performance in new deployment environments while fine-tuning could have been used to help remedy the issue. Similarly, changes in patient population and disease spectrum induced by the COVID-19 pandemic would have triggered a proactive reassessment of model performance under the proposed framework. Additionally, time-resolved assessment of model performance might have helped identify issues related to prediction timeliness. In short, the proposed framework provides a mechanism through which adequacy of performance can be robustly monitored and deviations can potentially be remedied, while also allowing dynamic compensation for institutional and operational shift.

**Future challenges**

Beyond the algorithmic performance of prognostic and diagnostic AI-MDs, human factors and practical usage significantly influence the clinical utility of these devices. Early in the adoption process at a particular institution, these factors are particularly intertwined and difficult to separate. For diagnostic AI-MDs as well as prognostic AI-MDs during the pre-deployment phase, silent evaluation can be employed to assess algorithmic performance independently. However, once a prognostic device is in widespread use, its predictions (ideally) start to influence clinicians' actions, complicating the assessment of purely algorithmic performance. The issue is then further complicated by the fact that the state of the clinical adoption process itself can substantially impact the practical utility of the device at any given time.

This issue is touched on in the DECIDE-AI guideline, which recognizes the importance of human factors considerations in the early stage evaluation of decision support systems.[17] Going further, van de Sande et al. recommend that specific evaluation tools be developed to allow robust evaluation of implementation outcomes.[32] In any case, human factors should be considered during the deployment and continued operation of prognostic and diagnostic AI-MDs. Human factors evaluations should be conducted according to appropriate guidelines, as these become available.

Finally, a potential challenge identified in the proposed framework exists regarding possible feedback during fine-tuning of prognostic AI-MDsTo be more specific, consider a prognostic model designed to predict the likelihood of a future adverse outcome. Upon deployment, this model ideally aids clinicians in delivering optimal care, thereby mitigating the risk of the anticipated adverse outcome. Should repeated model fine-tuning become necessary, e.g. due to operational shift, it is ideal to utilize the latest data, which inherently reflects this shift. However, this entails that the model is being fine-tuned on data that may have been shaped by the prognoses provided by earlier model versions. These changes can potentially alter the underlying prediction problem that the model aims to solve, and such alterations may or may not be compensable through fine-tuning. In the opinion of the authors, this feedback issue requires further examination to ensure that fine-tuning does not negatively affect model performance or device utility in unforeseen ways over the long term. Assessing the practical importance of this issue and, if necessary, finding viable solutions to it will require further research efforts by clinicians, researchers, and device manufacturers.

**Conclusion**

Recognizing the particular responsibility that comes with deploying AI models in a healthcare setting and considering that safety and effectiveness are paramount for any medical device, the deployment of prognostic and diagnostic AI-MDs should be conceptualized as a continuous process. Beyond initial installation, their deployment necessitates continuous oversight to consistently maintain high standards of performance. This oversight needs to extend beyond one-time validation and entail a clearly defined process for repeating validation and fine-tuning, particularly in deployment environments where operational shift can reasonably be expected to affect device performance. The presented viewpoint outlines a framework from which specific validation and fine-tuning strategies can be developed. By adopting repeating validation and fine-tuning, AI models in healthcare will hopefully become more effective and reliable, thus enhancing patient care and outcomes.

## Acknowledgments

ChatGPT using the GPT-4 architecture (OpenAI, San Francisco, USA) was used to support language editing for this work.

## Contributors

## Declaration of interests

The authors declare the following competing interests: All authors are employees of and/or hold equity in x-cardiac GmbH (Berlin, Germany). AM receives partial support through funding from the DHZB Foundation (Berlin, Germany), which holds equity in x-cardiac GmbH. FH, KB, and AM are named as inventors on a patent application, submitted by x-cardiac GmbH. It covers the application of elements of the presented framework for a specific class of prognostic use cases.

| Aspect | Traditional External Validation plus Vigilance and Post-Market Surveillance | Repeating Local Validation and Fine-Tuning |
|---|---|---|
| **Validation Timing** | Pre-deployment only, followed by passive monitoring | Continuous throughout the product life cycle |
| **Data Utilization** | Primarily based on pre-deployment data; limited real-world data integration | Extensive use of real-world data for ongoing adjustments |
| **Adaptability** | Low; updates and adaptations are infrequent | High; enables rapid response to institutional and operational shift |
| **Stakeholder Engagement** | Primarily manufacturer-driven; limited ongoing clinician involvement | Active involvement from clinicians |
| **Regulatory Compliance** | Compliance achieved pre-deployment; ongoing compliance monitored passively | Dynamic compliance with evolving state-of-the-art methods |
| **Risk Management** | Reactive; based on reported incidents or adverse events | Proactive; anticipates and mitigates risks based on performance data |
| **Resource Requirements** | Significant for initial validation; variable for ongoing vigilance and post-market surveillance | Potentially higher due to ongoing validation but can lead to greater product efficiency and effectiveness |

**Table 1:** Comparison of traditional external validation plus vigilance and post-market surveillance versus REVAFT. The REVAFT framework emphasizes a continuous, dynamic process throughout the product life cycle, leveraging real-world data and stakeholder involvement for improved outcomes and compliance.

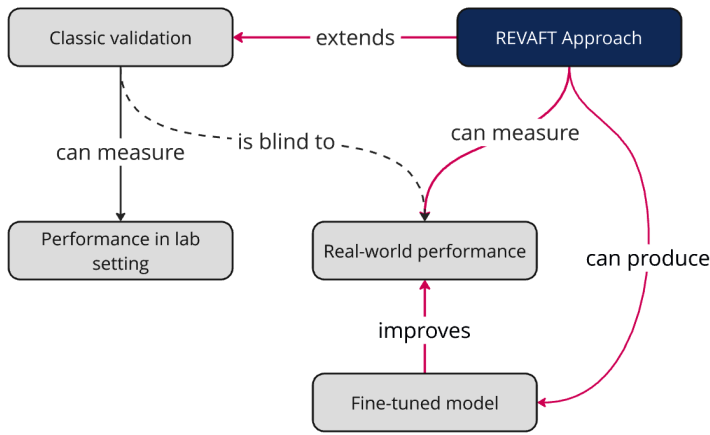

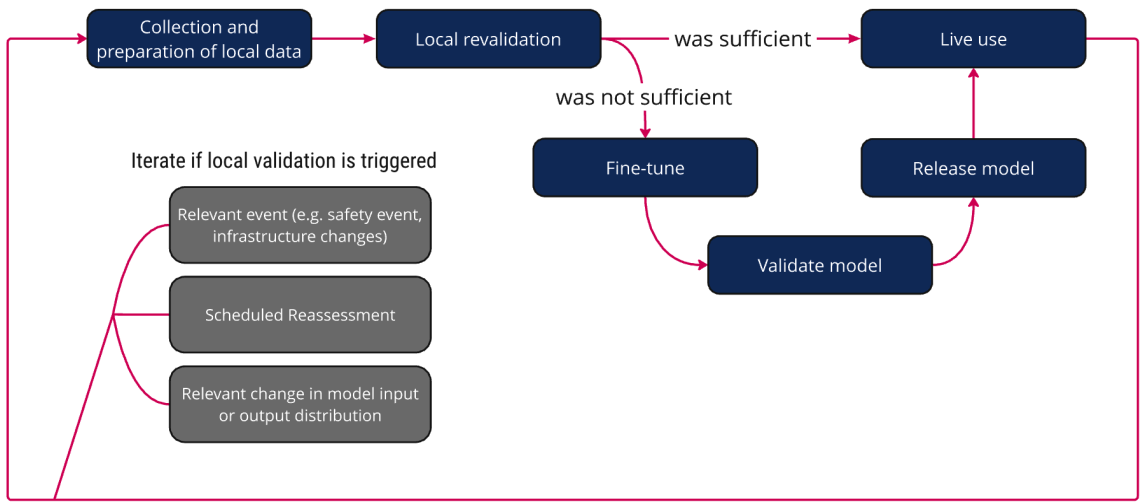

**Figure 1:** Illustration of the traditional validation process in contrast to the dynamic, real-world focused REVAFT (REpeating VAlidation and Fine-Tuning) framework for prognostic and diagnostic AI-MDs. Panel A depicts the expansion from classic validation, limited to lab settings, to the REVAFT approach that incorporates real-world performance data and includes model enhancement through fine-tuning. Panel B details the repeating aspects of the REVAFT framework, highlighting the cyclical process of local data collection, revalidation, and if necessary, fine-tuning.